\documentclass{article} %
\usepackage{iclr2027_conference,times}
\usepackage{comment}
\usepackage{multirow}
\usepackage{amsthm}
\usepackage[utf8]{inputenc} %
\usepackage[T1]{fontenc}    %

\usepackage{amsmath,amsfonts,bm}

\def\eqref#1{equation~\ref{#1}}

\def\1{\bm{1}}

\DeclareMathAlphabet{\mathsfit}{\encodingdefault}{\sfdefault}{m}{sl}
\SetMathAlphabet{\mathsfit}{bold}{\encodingdefault}{\sfdefault}{bx}{n}

\usepackage{booktabs}       %
\usepackage{amsfonts}       %
\usepackage{nicefrac}       %
\usepackage{microtype}      %
\usepackage{xcolor}         %
\usepackage{float}
\usepackage{colortbl}
\usepackage{amsmath}
\usepackage{graphicx}
\usepackage{caption}
\usepackage{subcaption}
\usepackage{bbm}
\usepackage{enumitem}
\usepackage{comment}
\usepackage{makecell}
\usepackage{amssymb}
\usepackage{hyperref}
\usepackage{url}
\usepackage{adjustbox}
\usepackage{pifont}
\usepackage{placeins}
\usepackage{amssymb}

\newcommand{\alg}{\textsc{Hawk}}

\title{HAWK: Rethinking Multimodal Drafting for Speculative Decoding}

\author{
  Wenhan Yang$^{1}$ \quad Anirudh Rao$^{2}$ \quad Ashwin Chandra$^{2}$ \\
  {\normalfont $^{1}$University of California, Los Angeles \quad $^{2}$Samsung Research America} \\
  {\normalfont\texttt{hangeryang18@g.ucla.edu}}
}

\iclrfinalcopy

\begin{document}

\maketitle
\begin{abstract}
    Speculative decoding has achieved substantial lossless speedups for LLMs, but remains less effective for large vision-language models (LVLMs), where lightweight drafters struggle to use rich multimodal information. A second limitation is that standard distillation supervises the drafter only along the original training trajectory, without modeling how target predictions shift after the drafter's own proposals. As drafting moves away from this trajectory, the drafter can increasingly disagree with the target, reducing acceptance in later steps.
We propose HAWK to address both limitations. HAWK uses representation similarity to select informative target layers and learns how to combine their hidden states. For visual information, it directly provides the drafter with compressed visual hidden states from the target model instead of raw visual tokens, making the visual information easier for a shallow drafter to use. HAWK also trains the drafter to capture how target predictions change after its own proposals, improving its agreement with the target during multi-step drafting. On SmolVLM-256M across ten multimodal benchmarks, HAWK raises average acceptance length from 3.32 to 4.08 and speedup from $2.19\times$ to $2.60\times$ over EAGLE-3 under greedy decoding, and from 2.89 to 3.41 and $1.92\times$ to $2.19\times$ under sampling.

\end{abstract}
\section{Introduction}

Large language models (LLMs) achieve strong performance across many tasks, but autoregressive generation remains expensive because each output token requires a sequential model forward pass~\citep{ouyang2022training,deepseek2025r1,leviathan2023fast}. Speculative decoding reduces this cost by using a lightweight drafter to propose multiple tokens that are verified in parallel by the target model~\citep{leviathan2023fast,chen2023accelerating}. Recent methods improve the trade-off between drafting cost and target agreement through auxiliary prediction heads~\citep{cai2024medusa}, early-exit drafting~\citep{elhoushi2024layerskip,liu2024kangaroo}, or lightweight drafters conditioned on target hidden states~\citep{li2024eagle,li2024eagle2,li2025eagle3}.

Extending speculative decoding to large vision-language models (LVLMs) is more difficult because lightweight drafters must also use rich visual context. Existing methods therefore reduce the visual input through token compression, adaptive selection, or by reusing visual information already encoded by the target~\citep{kang2026vispec,hu2026dream,huang2025specvlm,xie2026hivis}. However, reducing the number of visual tokens does not necessarily make their information easier for a shallow drafter to use, and aggressive compression may discard useful information. \alg{} instead gives the drafter the visual hidden states the target computes during prefill, so the drafter starts from features the target has already built rather than from raw image tokens.

A second limitation lies in draft-model training. Standard distillation supervises predictions along the original response trajectory, while multi-step drafting conditions on the drafter's own proposals. Once a proposed token changes the context, the target model's subsequent prediction may also change. Existing approaches improve supervision at such states by matching target logits or features, including states induced by the drafter~\citep{li2025eagle3,zhang2025hass,zhou2024distillspec,lei2026draft}. We instead supervise \emph{how} the target prediction changes relative to the original trajectory. Since standard distillation already teaches the target distribution on the original context, this objective targets only what changes once the drafter's token enters the context.

Based on these observations, we propose \alg{}, a multimodal speculative decoding framework that improves both the information provided to the drafter and the supervision used to train it. For target hidden states, rather than selecting layers only by depth, \alg{} uses inter-layer representation similarity to identify distinct regions, selects representative layers within each region, and learns how to combine their features. For visual information, it reuses and compresses target visual hidden states computed during prefill, providing richer visual conditioning without requiring the drafter to recover the same information from raw visual tokens. During training, \alg{} additionally substitutes the drafter's own proposals into the context and trains the drafter to match the resulting shift in target predictions.

Across ten multimodal benchmarks on SmolVLM-256M, \alg{} consistently improves speculative drafting. Under greedy decoding, it increases average acceptance length from 3.32 to 4.08 and speedup ratio from $2.19\times$ to $2.60\times$ over EAGLE-3, while also outperforming ViSpec and HiViS on average acceptance length and speedup. Under sampling, it improves average acceptance length from 2.89 to 3.41 and speedup ratio from $1.92\times$ to $2.19\times$ over EAGLE-3. On the larger LLaVA-1.6-7B and Qwen2.5-VL-7B targets, \alg{} raises average acceptance length under greedy decoding from 3.73 to 4.02 and from 3.19 to 3.40, and improves both acceptance length and speedup on every benchmark at both temperatures.

Our contributions are twofold:
\begin{itemize}
    \item We improve multimodal drafting inputs through representation-based target-layer selection, learned feature fusion, and compressed visual hidden states reused from target prefill.
    \item We complement standard distillation with supervision on target prediction shifts caused by the drafter's own proposals, improving training for multi-step drafting without additional inference-time computation.
\end{itemize}

\section{Related Work}

\paragraph{Speculative decoding for language models.}
Speculative decoding verifies several drafted tokens in one target forward pass~\citep{leviathan2023fast,chen2023accelerating}. Later methods differ in where those drafts come from. Medusa attaches extra decoding heads to the target~\citep{cai2024medusa}. LayerSkip and Kangaroo draft by exiting or reusing the target's own early layers, and LEAP searches over which layers to reuse~\citep{elhoushi2024layerskip,liu2024kangaroo,zhengleap}. EAGLE instead trains a separate lightweight drafter that reads target hidden states, with later versions adding dynamic draft trees and features from several layers~\citep{li2024eagle,li2024eagle2,li2025eagle3}. On the training side, EAGLE-3's training-time test feeds the drafter its own features during training~\citep{li2025eagle3}, and HASS aligns training with the drafter's own contexts~\citep{zhang2025hass}; in both, the input tokens remain those of the training sequence. DistillSpec distills the target distribution on drafter-generated sequences~\citep{zhou2024distillspec}, and Draft-OPD applies on-policy distillation to speculative draft models~\citep{lei2026draft}. \alg{} keeps the EAGLE-3 drafter and instead supervises how the target prediction changes after a draft token alters the context.

\paragraph{Speculative decoding for vision-language models.}
Vision-language speculative decoding must provide visual context to the drafter without incurring the full cost of the target model. Early approaches construct multimodal drafters using target vision encoders or specialized multimodal training~\citep{ganesan2025massv,lin2025speculative}. More recent methods increasingly reuse target-model information: DREAM injects intermediate target representations, while SpecVLM conditions an EAGLE-style drafter on target features together with adaptive visual compression~\citep{hu2026dream,huang2025specvlm}. ViSpec learns compact visual representations for a shallow drafter, whereas HiViS avoids explicit visual-token processing by transferring target representations in which visual and textual information have already been fused~\citep{kang2026vispec,xie2026hivis}. \alg{} also builds on target representations, and differs in how it chooses and fuses target layers and in passing pooled target visual states directly to the drafter.

\paragraph{Layer similarity.}
Representation-similarity measures such as CKA~\citep{kornblith2019similarity}
show that adjacent layers of deep networks are often highly similar, and
layer-pruning work uses this redundancy to remove layers of
LLMs~\citep{men2024shortgpt,gromov2024unreasonable}. HAWK uses the same signal
to choose which target layers a drafter reads.

\section{Method}
\label{sec:method}

\begin{figure}[t]
    \centering
    \includegraphics[width=\linewidth]{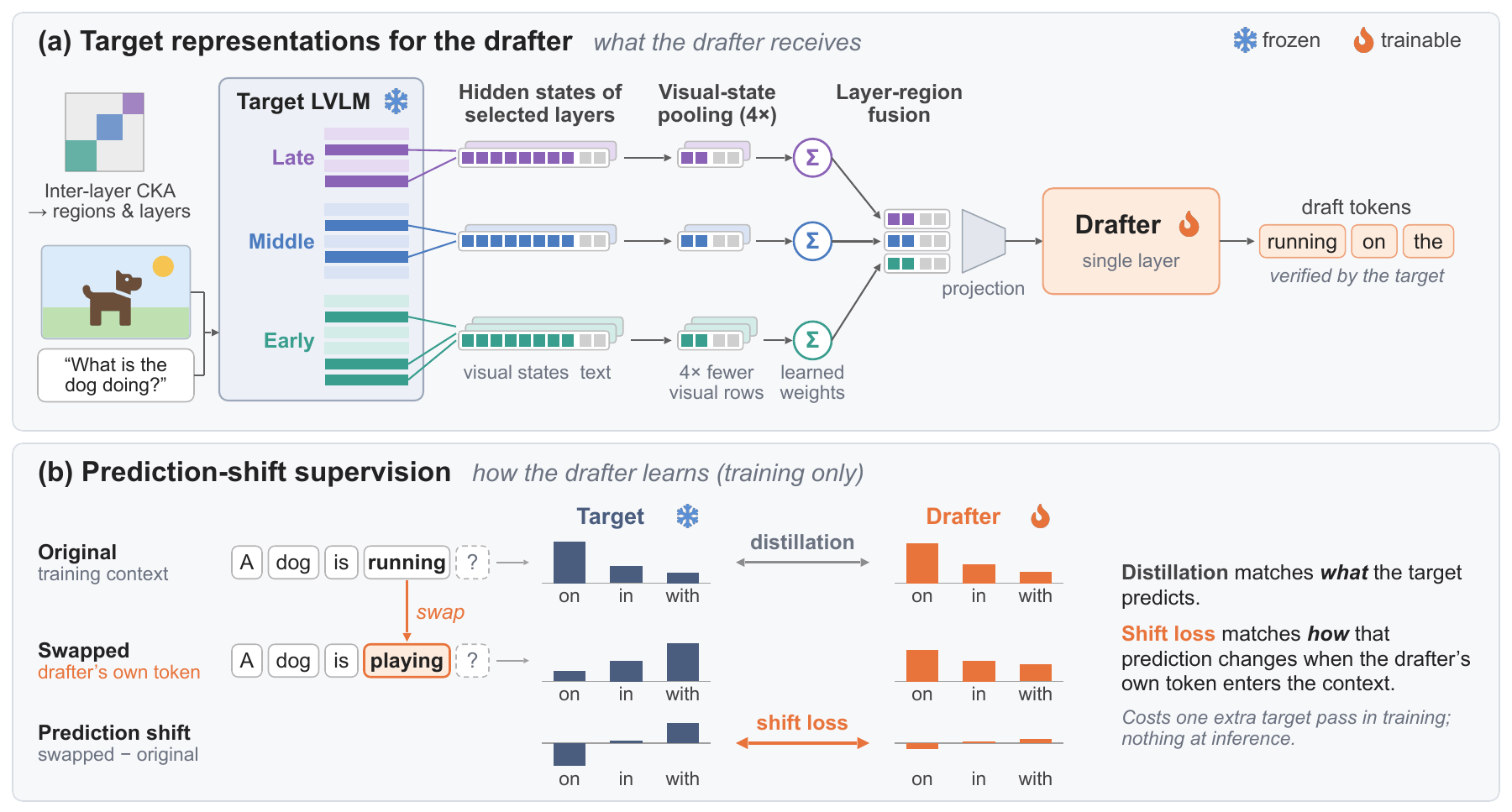}
    \caption{Overview of HAWK.
    (a) Target representations passed to the drafter.
    Inter-layer CKA divides the frozen target into early, middle, and late regions and selects representative layers.
    Visual hidden states from those layers are pooled by $4\times$, then combined within each region by learned weights.
    The three fused streams are projected into a single-layer drafter, whose proposals the target verifies.
    (b) Prediction-shift supervision, used only during training.
    Standard distillation matches the target's next-token prediction on the original context.
    The shift loss matches how that prediction changes after the drafter's own token replaces the original token.}
    \label{fig:hawk-framework}
\end{figure}

\alg{} changes two parts of an EAGLE-3 drafter: the target information it
receives (Sections~\ref{sec:fusion} and~\ref{sec:pool}) and how it is trained
(Section~\ref{sec:shift}). Figure~\ref{fig:hawk-framework} summarizes both.

\subsection{Layer-Region Fusion}
\label{sec:fusion}

EAGLE-3 combines hidden states from three target layers at fixed depths,
$(2, N/2, N-3)$ in 1-based indexing~\citep{li2025eagle3}. For SmolVLM-256M these
are decoder blocks $[1,14,26]$ in the 0-based indexing we use
(Appendix~\ref{app:impl}). Fixed positions need not match where the target
representations change, and reading more layers does not necessarily help,
because nearby layers are often redundant. We therefore split the target into
regions by representation similarity, select a few layers in each region, and
learn how to combine them.

\begin{figure}[t]
    \centering
    \begin{subfigure}[t]{0.32\linewidth}
        \centering
        \includegraphics[width=\linewidth]
            {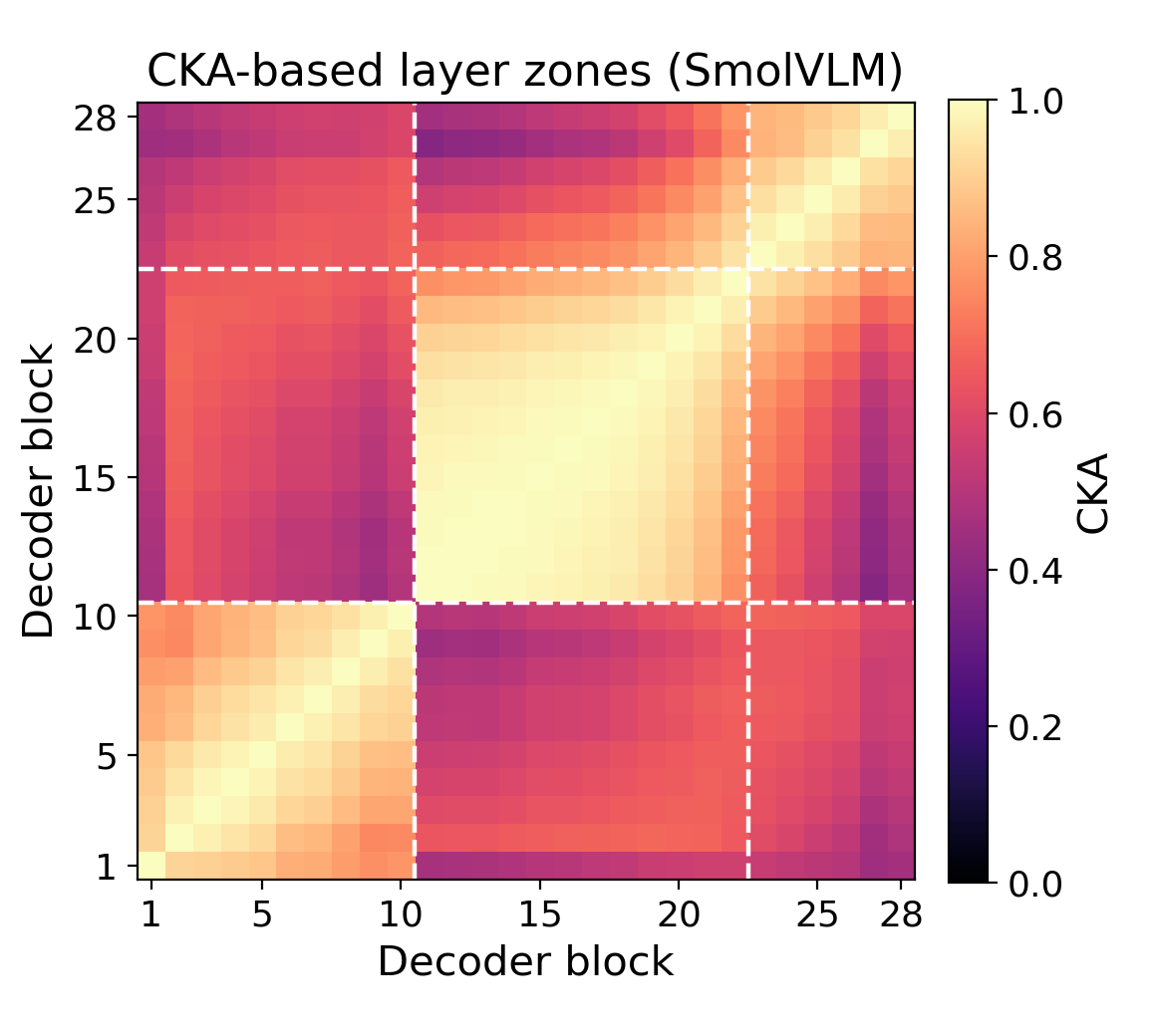}
        \caption{SmolVLM-256M.}
        \label{fig:layer-band-zone-smolvlm}
    \end{subfigure}
    \hfill
    \begin{subfigure}[t]{0.32\linewidth}
        \centering
        \includegraphics[width=\linewidth]
            {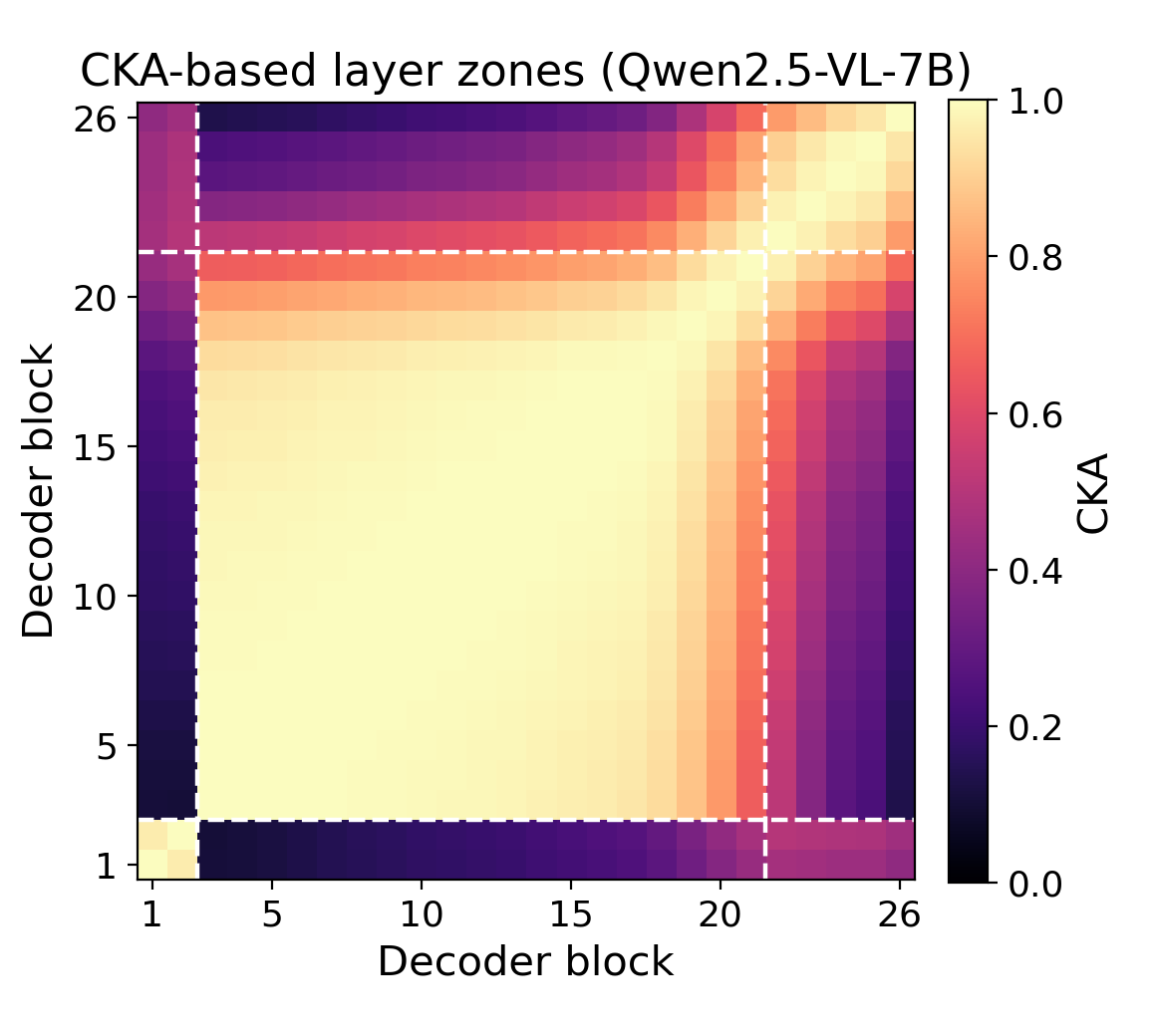}
        \caption{Qwen2.5-VL-7B.}
        \label{fig:layer-band-zone-qwen}
    \end{subfigure}
    \hfill
    \begin{subfigure}[t]{0.32\linewidth}
        \centering
        \includegraphics[width=\linewidth]
            {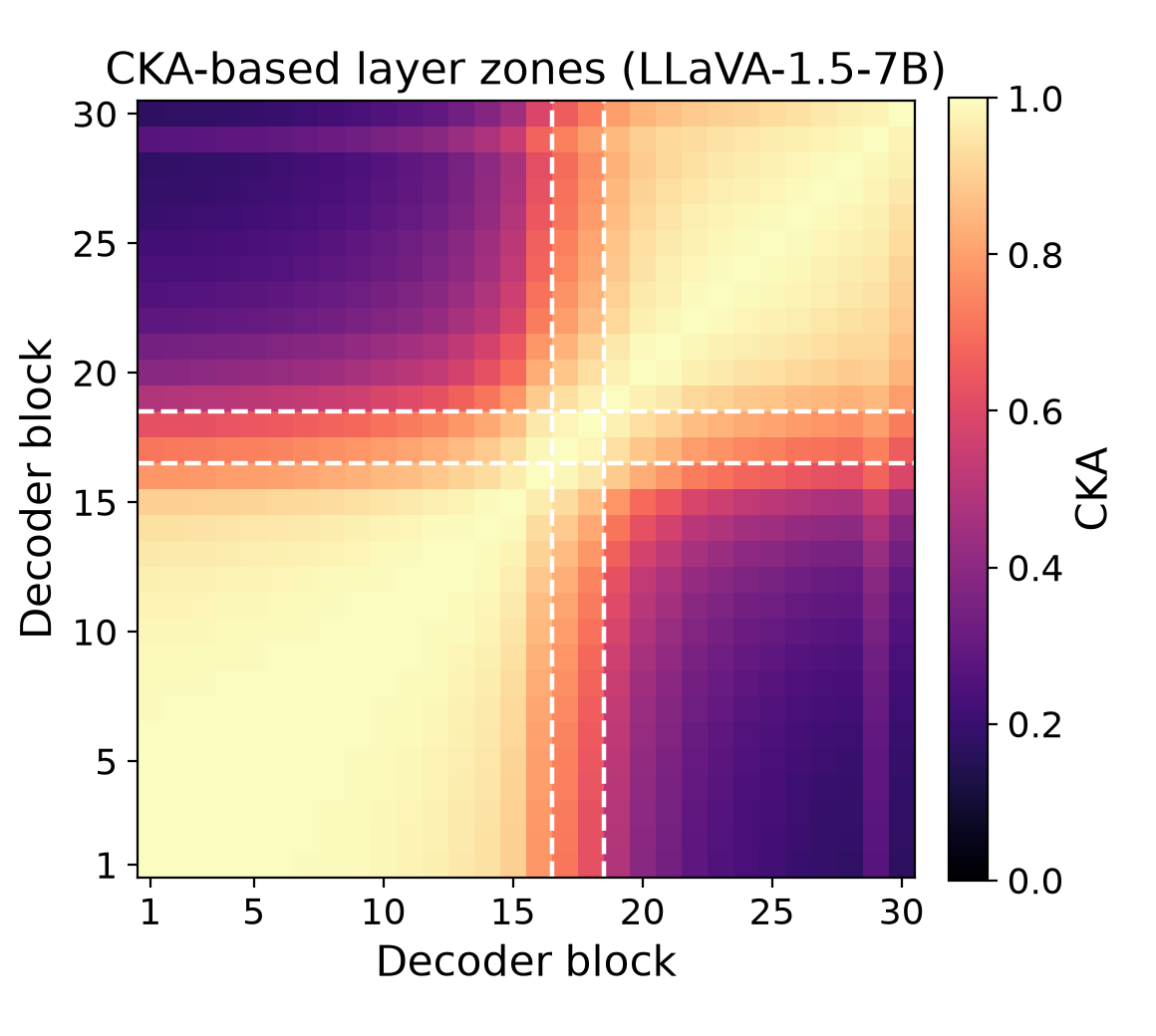}
        \caption{LLaVA-1.6-7B.}
        \label{fig:layer-band-zone-llava}
    \end{subfigure}
    \caption{Inter-layer CKA and the selected early, middle, and late regions.
    The boundaries differ across models, so one fixed depth rule does not fit all three.}
    \label{fig:layer-band-zone}
\end{figure}

\textbf{Region and layer selection.}
We measure layer similarity with linear CKA~\citep{kornblith2019similarity}.
For hidden states $X\in\mathbb{R}^{n\times d_X}$ and $Y\in\mathbb{R}^{n\times d_Y}$
of two layers at the same $n$ token positions, with columns centered,
\begin{equation}
\mathrm{CKA}(X,Y)=\frac{\|X^{\top}Y\|_F^2}{\|X^{\top}X\|_F\,\|Y^{\top}Y\|_F}.
\end{equation}
We compute CKA on the text positions of 48 training conversations, after
removing 32 outlier hidden dimensions (Appendix~\ref{app:cka}). An exhaustive
search over two boundaries then splits the target into three contiguous regions
whose layers are more similar to each other than to layers in other regions.
Within region $\mathcal{R}_r$, the CKA coverage of a subset
$\mathcal{S}_r\subseteq\mathcal{R}_r$ is
\begin{equation}
C_{\mathrm{CKA}}(\mathcal{S}_r)=\frac{1}{|\mathcal{R}_r|}\sum_{l\in\mathcal{R}_r}\max_{s\in\mathcal{S}_r}\mathrm{CKA}(l,s).
\end{equation}
Starting from an empty set, we add the layer with the largest coverage gain
until the gain falls below $\epsilon_{\mathrm{CKA}}=0.008$.
Figure~\ref{fig:layer-band-zone} shows the resulting regions;
Appendix~\ref{app:cka} gives the coverage curves and selected layers.

\textbf{Learnable fusion.}
Each region feeds one of the three EAGLE-3 input streams. We mix its selected
layers with learned softmax weights,
\begin{equation}
\bar{\mathbf{H}}_r=\sum_{s\in\mathcal{S}_r}\alpha_{r,s}\tilde{\mathbf{H}}^{(s)},
\qquad
\alpha_{r,s}=\frac{\exp(a_{r,s})}{\sum_{j\in\mathcal{S}_r}\exp(a_{r,j})},
\end{equation}
where $a_{r,s}$ are learnable scalars and $\tilde{\mathbf{H}}^{(s)}$ is the pooled hidden-state sequence of layer
$s$ (Section~\ref{sec:pool}). We then apply the EAGLE-3 stream normalization
$\mathcal{N}_r$ and input projection $P$,
\begin{equation}
\mathbf{U}=P\big([\mathcal{N}_{\mathrm{early}}(\bar{\mathbf{H}}_{\mathrm{early}});\,\mathcal{N}_{\mathrm{middle}}(\bar{\mathbf{H}}_{\mathrm{middle}});\,\mathcal{N}_{\mathrm{late}}(\bar{\mathbf{H}}_{\mathrm{late}})]\big),
\end{equation}
so the drafter keeps the EAGLE-3 input interface. Figure~\ref{fig:app-alpha}
shows the learned weights.

\subsection{Visual State Pooling}
\label{sec:pool}

Recent multimodal drafters shorten the visual input with learned compression or
drop visual tokens altogether~\citep{kang2026vispec,xie2026hivis}. We instead
give the drafter the visual hidden states that the target has already computed
during prefill, and pool them to keep the sequence short. At each selected
layer $l$, we split every contiguous run of image rows (one 64-row tile in
SmolVLM) into consecutive groups $\mathcal{G}_j$ of $c$ rows and average each
group:
\begin{equation}
\tilde{\mathbf{h}}^{(l)}_j=\frac{1}{|\mathcal{G}_j|}\sum_{i\in\mathcal{G}_j}\mathbf{h}^{(l)}_i.
\end{equation}
All layers use the same groups, text rows are left unchanged, and each pooled
row keeps the position id of the first row in its group
(Appendix~\ref{app:impl}). We use $c=4$. Moderate pooling improves acceptance,
while stronger pooling removes useful information (Section~\ref{sec:ablation},
Appendix~\ref{app:compression}). Pooling changes only the drafter's input; the
target still processes the full visual sequence during prefill and
verification.

\subsection{Prediction-Shift Supervision}
\label{sec:shift}

Standard distillation trains the drafter to match target predictions along the
training sequence. In multi-step drafting, however, the drafter conditions on
its own proposals, and even a plausible alternative token can change what the
target predicts next. We therefore train the drafter so that the change in its
prediction after such a substitution matches the change in the target's
prediction.

\textbf{Alternative token selection.}
Let $x_t$ be the response token at position $t$. At the first rollout step of EAGLE-3's training-time test (one unrolled draft step), the
drafter proposes $\hat{x}_t=\arg\max_x p_D(x\mid x_{<t})$. We keep the positions
where this proposal differs from the target's top-1 prediction:
\begin{equation}
\mathcal{S}=\big\{t:\hat{x}_t\neq\arg\max_x p_T(x\mid x_{<t})\big\}.
\end{equation}
For each $t\in\mathcal{S}$, the drafter runs its second rollout step with
$\hat{x}_t$ in place of $x_t$. Its own output feature from the first step and
the prefix cache stay the same; only the input token changes. EAGLE-3's
training-time test trains the drafter on its own feature drift but never on a
context whose token differs from the training sequence; the substitution
supervises exactly this token pathway, at the same rollout state
(Appendix~\ref{app:shift}).

\textbf{Logit-shift objective.}
Let $z^{\mathrm{orig}}_{T,t}$ and $z^{\mathrm{orig}}_{D,t}$ be the target and
drafter logits for the token following position $t$ under the original context, and $z^{\mathrm{alt}}_{T,t}$ and
$z^{\mathrm{alt}}_{D,t}$ the logits after substitution. We map the target logits
to the draft vocabulary $V_D$ and center every logit vector,
$\mathrm{center}(z)=z-\mathrm{mean}(z)$, which removes offsets that softmax
ignores. The target and drafter shifts are
\begin{equation}
\Delta_{T,t}=\mathrm{center}(z^{\mathrm{alt}}_{T,t})-\mathrm{center}(z^{\mathrm{orig}}_{T,t}),
\qquad
\Delta_{D,t}=\mathrm{center}(z^{\mathrm{alt}}_{D,t})-\mathrm{center}(z^{\mathrm{orig}}_{D,t}),
\end{equation}
and the training objective is
\begin{equation}
\mathcal{L}=\mathcal{L}_{\mathrm{std}}+\lambda_{\mathrm{shift}}\mathcal{L}_{\mathrm{shift}},
\qquad
\mathcal{L}_{\mathrm{shift}}=\operatorname*{mean}_{t\in\mathcal{S}}\frac{1}{|V_D|}\big\|\Delta_{D,t}-\Delta_{T,t}\big\|_2^2,
\qquad
\lambda_{\mathrm{shift}}=0.1,
\end{equation}
where $\mathcal{L}_{\mathrm{std}}$ is the standard distillation loss. Because
centered logits equal log-probabilities up to a shared constant,
$\Delta_{T,t}$ is the centered log-ratio
$\log p^{\mathrm{alt}}_T-\log p^{\mathrm{orig}}_T$. $\mathcal{L}_{\mathrm{std}}$
fits the level of the prediction, and $\mathcal{L}_{\mathrm{shift}}$ fits how
it changes (Appendix~\ref{app:shift}). As a function of the input token,
$\Delta_{T,t}$ is a finite difference of the target's log-probabilities along
$x_t\to\hat{x}_t$, so $\mathcal{L}_{\mathrm{shift}}$ matches the drafter's
sensitivity to its input token to that of the target.

\textbf{Batched target supervision.}
Scoring each substitution separately would need one target forward pass per
selected position. We instead substitute all positions in $\mathcal{S}$ into one
copy of the sequence and obtain $z^{\mathrm{alt}}_{T}$ with a single extra
target pass. The original logits $z^{\mathrm{orig}}_{T}$ and the target hidden
states come from the offline cache. The extra pass runs only during training and adds
no inference cost.

\section{Experiments}

\subsection{Experimental Setup}
\label{sec:setup}

\textbf{Models and benchmarks.}
We use SmolVLM-256M-Instruct~\citep{marafioti2025smolvlm} as the main target and
report LLaVA-1.6-7B and Qwen2.5-VL-7B in Section~\ref{sec:7b}. We evaluate on ten
benchmarks: MMMU, SEED-Bench, MME, VQAv2, and MM-Vet for general understanding;
MathVista, ScienceQA, and GQA for visual reasoning; and TextVQA and DocVQA for
OCR~\citep{yue2023mmmu,li2023seed,fu2026mme,goyal2017making,yu2023mmvet,lu2024mathvista,lu2022learn,hudson2019gqa,singh2019towards,mathew2020docvqa}.
Each benchmark uses 80 prompts (Appendix~\ref{app:bench}).

\textbf{Baselines.}
We compare with EAGLE-3~\citep{li2025eagle3}, trained with the same data,
offline cache, and budget as \alg{}, and with ViSpec~\citep{kang2026vispec} and
HiViS~\citep{xie2026hivis}, trained with their official implementations. All
methods use the same evaluator, draft-tree configuration, and prompts
(Appendix~\ref{app:baselines}).

\textbf{Metrics.}
Following HiViS, we report the speedup ratio (SR), the generation token rate
relative to autoregressive decoding, and the average acceptance length $\tau$,
the number of tokens produced per verification step
(Appendix~\ref{app:timing}). We do not report task accuracy, since speculative
decoding leaves the target's output distribution unchanged.

\textbf{Training setup.}
Following HiViS, we train on 68K text-only conversations from
ShareGPT and 68K multimodal conversations from
LLaVA-Mix665K~\citep{sharegpt2023,liu2023visual}, with responses regenerated by
the target. The EAGLE-3 and \alg{} drafters have a single decoder layer and are
trained for 20 epochs (Appendix~\ref{app:impl}).

\textbf{Inference setup.}
Following ViSpec and HiViS, we evaluate greedy decoding ($T=0$) and sampling
($T=1$). All runs use one NVIDIA RTX A5000 GPU with batch size 1. The draft tree
has depth 5, top-10 expansion, and 60 candidate tokens
(Appendix~\ref{app:timing}).

\subsection{Main Results}

\begin{table}[t]
\caption{Speedup ratio (SR) and average acceptance length ($\tau$) on SmolVLM-256M under greedy decoding ($T=0$, top) and sampling ($T=1$, bottom). Averages at both temperatures cover ten benchmarks.}
\label{tab:main-results}
\label{tab:sampling-results}
\centering
\normalsize
\renewcommand{\arraystretch}{1.2}
\begin{adjustbox}{max width=\linewidth}
\begin{tabular}{@{}l@{\hspace{4pt}}l@{\hspace{6pt}}r@{\hspace{4pt}}r%
@{\hspace{7pt}}r@{\hspace{4pt}}r@{\hspace{7pt}}r@{\hspace{4pt}}r%
@{\hspace{7pt}}r@{\hspace{4pt}}r@{\hspace{7pt}}r@{\hspace{4pt}}r%
@{\hspace{7pt}}r@{\hspace{4pt}}r@{\hspace{7pt}}r@{\hspace{4pt}}r%
@{\hspace{7pt}}r@{\hspace{4pt}}r@{\hspace{7pt}}r@{\hspace{4pt}}r%
@{\hspace{7pt}}r@{\hspace{4pt}}r@{\hspace{7pt}}r@{\hspace{4pt}}r@{}}
\toprule
 &  & \multicolumn{10}{c}{General} & \multicolumn{6}{c}{Reasoning} &
 \multicolumn{4}{c}{OCR} & \multicolumn{2}{c}{} \\
\cmidrule(lr){3-12} \cmidrule(lr){13-18} \cmidrule(lr){19-22}
\multirow{2}{*}{Model} & \multirow{2}{*}{Methods} & \multicolumn{2}{c}{MMMU} &
\multicolumn{2}{c}{SEED-Bench} & \multicolumn{2}{c}{MME} & \multicolumn{2}{c}{VQAv2} &
\multicolumn{2}{c}{MM-Vet} & \multicolumn{2}{c}{MathVista} &
\multicolumn{2}{c}{SQA} & \multicolumn{2}{c}{GQA} & \multicolumn{2}{c}{TextVQA}
& \multicolumn{2}{c}{DocVQA} & \multicolumn{2}{c}{Avg} \\
 & & SR & $\tau$ & SR & $\tau$ & SR & $\tau$ & SR & $\tau$ & SR & $\tau$ & SR &
 $\tau$ & SR & $\tau$ & SR & $\tau$ & SR & $\tau$ & SR & $\tau$ & SR & $\tau$ \\
\midrule
\multicolumn{24}{c}{\textbf{Greedy decoding ($T=0$)}} \\
\midrule
\multirow{4}{*}{\makecell[l]{SmolVLM\\256M}}
 & EAGLE-3 & 1.99 & 2.87 & 2.42 & 4.29 & 2.27 & 3.87 & 2.45 & 4.16 & 2.15 & 3.40 & 2.44 & 2.99 & 1.94 & 2.75 & 2.54 & 4.28 & 2.11 & 2.56 & 1.61 & 2.00 & 2.19 & 3.32 \\
 & ViSpec & 1.62 & 2.24 & 2.34 & 4.38 & 2.06 & 3.56 & 2.35 & 4.48 & 2.04 & 3.47 & 2.38 & 3.32 & 1.84 & 2.74 & 2.34 & 4.16 & 2.27 & 3.02 & 1.57 & 2.00 & 2.08 & 3.34 \\
 & HiViS & 1.98 & 2.70 & 2.60 & 4.73 & 2.37 & 4.01 & 2.72 & 4.71 & 2.40 & 3.85 & 3.00 & 3.87 & 2.24 & 3.25 & 2.77 & 4.70 & 2.56 & 3.25 & 1.77 & 2.10 & 2.44 & 3.72 \\
 & HAWK & \textbf{2.23} & \textbf{3.39} & \textbf{2.74} & \textbf{5.02} & \textbf{2.55} & \textbf{4.54} & \textbf{2.73} & \textbf{4.80} & \textbf{2.42} & \textbf{4.04} & \textbf{3.24} & \textbf{4.26} & \textbf{2.35} & \textbf{3.54} & \textbf{2.82} & \textbf{4.90} & \textbf{2.85} & \textbf{3.63} & \textbf{2.03} & \textbf{2.71} & \textbf{2.60} & \textbf{4.08} \\
\midrule
\multicolumn{24}{c}{\textbf{Sampling ($T=1$)}} \\
\midrule
 &  & \multicolumn{10}{c}{General} & \multicolumn{6}{c}{Reasoning} &
 \multicolumn{4}{c}{OCR} & \multicolumn{2}{c}{} \\
\cmidrule(lr){3-12} \cmidrule(lr){13-18} \cmidrule(lr){19-22}
\multirow{2}{*}{Model} & \multirow{2}{*}{Methods} & \multicolumn{2}{c}{MMMU} &
\multicolumn{2}{c}{SEED-Bench} & \multicolumn{2}{c}{MME} & \multicolumn{2}{c}{VQAv2} &
\multicolumn{2}{c}{MM-Vet} & \multicolumn{2}{c}{MathVista} &
\multicolumn{2}{c}{SQA} & \multicolumn{2}{c}{GQA} & \multicolumn{2}{c}{TextVQA}
& \multicolumn{2}{c}{DocVQA} & \multicolumn{2}{c}{Avg} \\
 & & SR & $\tau$ & SR & $\tau$ & SR & $\tau$ & SR & $\tau$ & SR & $\tau$ & SR &
 $\tau$ & SR & $\tau$ & SR & $\tau$ & SR & $\tau$ & SR & $\tau$ & SR & $\tau$ \\
\midrule
\multirow{4}{*}{\makecell[l]{SmolVLM\\256M}}
 & EAGLE-3 & 1.65 & 2.42 & 2.26 & 3.88 & 2.03 & 3.28 & 2.20 & 3.67 & 1.66 & 2.64 & 2.04 & 2.79 & 1.68 & 2.35 & 2.15 & 3.48 & 2.05 & 2.48 & 1.45 & 1.93 & 1.92 & 2.89 \\
 & ViSpec & 1.42 & 1.99 & 2.19 & 4.00 & 1.85 & 2.99 & 2.22 & \textbf{4.17} & 1.75 & 2.83 & 1.98 & 3.05 & 1.71 & 2.53 & 2.05 & 3.63 & 2.09 & 2.88 & 1.42 & 1.95 & 1.87 & 3.00 \\
 & HiViS & 1.65 & 2.35 & \textbf{2.47} & \textbf{4.38} & 2.01 & 3.34 & 2.32 & 3.76 & \textbf{1.92} & \textbf{3.13} & 2.37 & 3.37 & 1.90 & 2.82 & 2.35 & 3.95 & 2.35 & 3.10 & 1.58 & 2.06 & 2.09 & 3.23 \\
 & HAWK & \textbf{1.72} & \textbf{2.68} & 2.43 & 4.27 & \textbf{2.14} & \textbf{3.48} & \textbf{2.44} & 4.09 & 1.91 & 3.03 & \textbf{2.51} & \textbf{3.73} & \textbf{1.96} & \textbf{3.01} & \textbf{2.39} & \textbf{3.99} & \textbf{2.63} & \textbf{3.34} & \textbf{1.77} & \textbf{2.46} & \textbf{2.19} & \textbf{3.41} \\
\bottomrule
\end{tabular}
\end{adjustbox}
\end{table}

Table~\ref{tab:main-results} compares \alg{} with EAGLE-3, ViSpec, and HiViS on
SmolVLM-256M. \alg{} has the highest average speedup ratio and acceptance length
at both temperatures.

\textbf{Greedy decoding.}
At $T=0$, \alg{} has the highest SR and $\tau$ on all ten benchmarks. Its average
$\tau$ is 4.08, compared with 3.32 for EAGLE-3, 3.34 for ViSpec, and 3.72 for
HiViS, and its average SR is $2.60\times$, compared with $2.19\times$,
$2.08\times$, and $2.44\times$. The largest speedup is $3.24\times$ on
MathVista, where $\tau$ rises from 2.99 with EAGLE-3 to 4.26.

\textbf{Sampling.}
At $T=1$, \alg{} again has the best averages. Compared with EAGLE-3, it raises $\tau$ from 2.89 to 3.41 and SR from
$1.92\times$ to $2.19\times$, with gains on all ten benchmarks. It also exceeds
ViSpec (3.00, $1.87\times$) and HiViS (3.23, $2.09\times$) on average.

\textbf{Gains by task type.}
On SmolVLM-256M under greedy decoding,
$\tau$ increases by 14--19\% on the five general benchmarks and GQA, and by
29--42\% on MathVista, ScienceQA, TextVQA, and DocVQA. The margin over HiViS is
largest on MMMU (3.39 vs.\ 2.70) and DocVQA (2.71 vs.\ 2.10), where HiViS gains
little or nothing over EAGLE-3. The largest relative gains are on reasoning and OCR
benchmarks, whose answers depend most on visual detail.

\begin{figure}[t]
\centering
\includegraphics[width=0.75\linewidth]{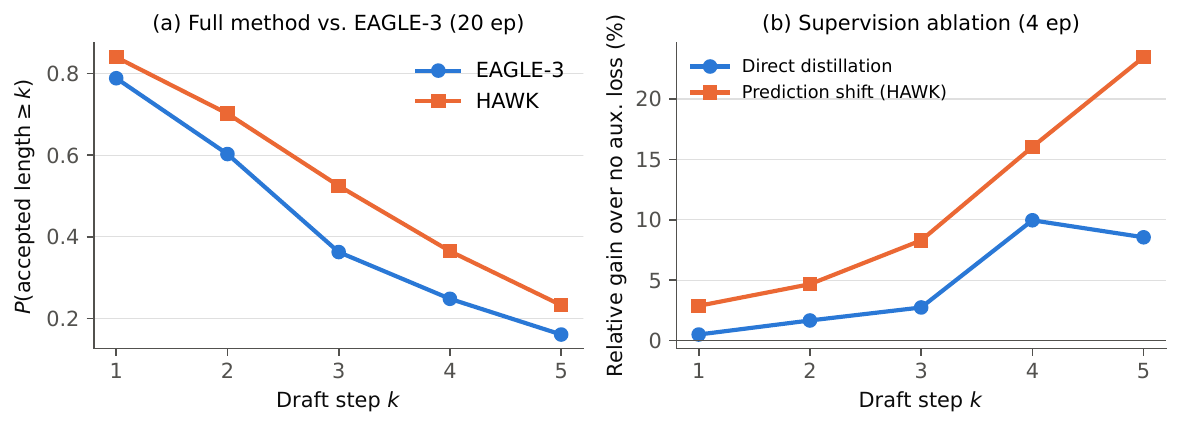}
\caption{(a) Probability that the first $k$ draft tokens are all accepted,
EAGLE-3 vs.\ \alg{}. (b) Relative change of the same quantity over the drafter
trained without auxiliary loss, for the supervision variants of
Table~\ref{tab:ablation-shift}. The gain from prediction-shift supervision
grows from 3\% at $k=1$ to 23\% at $k=5$, against at most 10\% for direct
distillation.}
\label{fig:app-per-step}
\end{figure}

\textbf{Acceptance by draft step.}
The gain over EAGLE-3 also grows with draft depth
(Figure~\ref{fig:app-per-step}a). The probability
that the first $k$ draft tokens are all accepted rises from 0.789 to 0.840 at
$k=1$ and from 0.161 to 0.233 at $k=5$, a relative gain of 6\% at the first step
and 45\% at the fifth. Isolating prediction-shift supervision in the four-epoch
ablation gives the same pattern: relative to the drafter trained without it, the
gain grows from 3\% at $k=1$ to 23\% at $k=5$, against at most 10\% for direct
distillation on the substituted contexts (Figure~\ref{fig:app-per-step}b). Later
steps are where the drafter conditions on its own proposals, which is the case
prediction-shift supervision is designed for.

\subsection{Ablation Studies}
\label{sec:ablation}

We ablate the main design choices of HAWK on SmolVLM-256M. All variants are trained for four epochs and evaluated under greedy decoding ($T=0$) on six benchmarks: SEED-Bench, MM-Vet, MathVista, ScienceQA, GQA, and DocVQA. We report acceptance length $\tau$ and speedup ratio (SR). Each study isolates one design choice while disabling the other HAWK components.

\textbf{Visual-state pooling.}
Without pooling, the drafter receives all visual hidden-state rows (832 for a 13-tile image) and achieves an average acceptance length of 3.05 with a $2.07\times$ speedup. As shown in Table~\ref{tab:ablation-pooling}, $4\times$ pooling reduces the sequence to 208 rows while improving both metrics to $\tau=3.10$ and $2.10\times$. More aggressive $16\times$ pooling leaves only 52 rows and reduces performance to $\tau=3.04$ and $2.07\times$, roughly returning to the no-pooling baseline. We therefore use $4\times$ pooling, which cuts the visual sequence by 75\% while improving both metrics.

\begin{table}[H]
\caption{Ablation of visual-state pooling on SmolVLM-256M. The other components are disabled.}
\label{tab:ablation-pooling}
\centering
\footnotesize
\setlength{\tabcolsep}{4.5pt}
\renewcommand{\arraystretch}{1.0}
\begin{tabular}{@{}lrrrrrrr@{}}
\toprule
Setting & SEED-Bench & MM-Vet & MathVista & SQA & GQA & DocVQA & Avg. \\
\midrule
\multicolumn{8}{@{}l}{\emph{Acceptance length $\tau$}} \\
No pooling & 4.03 & 3.15 & 2.85 & 2.36 & 3.80 & 2.08 & 3.05 \\
$4\times$ pooling & 4.03 & 3.25 & 2.92 & 2.45 & 3.86 & 2.09 & \textbf{3.10} \\
$16\times$ pooling & 4.00 & 3.20 & 2.75 & 2.43 & 3.82 & 2.03 & 3.04 \\
\midrule
\multicolumn{8}{@{}l}{\emph{Speedup ratio}} \\
No pooling & 2.36 & 2.04 & 2.18 & 1.74 & 2.28 & 1.79 & 2.07 \\
$4\times$ pooling & 2.39 & 2.10 & 2.21 & 1.81 & 2.31 & 1.77 & \textbf{2.10} \\
$16\times$ pooling & 2.38 & 2.07 & 2.18 & 1.81 & 2.29 & 1.67 & 2.07 \\
\bottomrule
\end{tabular}
\end{table}

\textbf{Prediction-shift supervision.}
For this study, we use the original target layers $[1,14,26]$ and disable visual-state pooling. Without the additional supervision, standard draft distillation gives $\tau=3.05$ and a $2.07\times$ speedup. Applying prediction-shift supervision with a larger weight of $0.3$ gives a similar result, reaching $\tau=3.11$ and $2.05\times$ speedup. In contrast, HAWK's prediction-shift supervision improves both metrics, reaching $\tau=3.17$ and $2.09\times$ speedup (Table~\ref{tab:ablation-shift}).

Direct distillation (Table~\ref{tab:ablation-shift}) uses the same substituted contexts, token selection, loss weight, and extra target pass, and differs only in the loss: cross-entropy to $p^{\mathrm{alt}}_T$ instead of squared error on the shift. It raises $\tau$ by 0.05 against 0.12 for the shift loss, and its speedup drops to $2.05\times$.

\begin{table}[H]
\caption{Ablation of prediction-shift supervision on SmolVLM-256M. The other components are disabled. Direct distillation matches the target distribution on substituted contexts, while HAWK supervises the induced prediction shift.}
\label{tab:ablation-shift}
\centering
\footnotesize
\setlength{\tabcolsep}{4.5pt}
\renewcommand{\arraystretch}{1.0}
\begin{tabular}{@{}lrrrrrrr@{}}
\toprule
Setting & SEED-Bench & MM-Vet & MathVista & SQA & GQA & DocVQA & Avg. \\
\midrule
\multicolumn{8}{@{}l}{\emph{Acceptance length $\tau$}} \\
No auxiliary loss & 4.03 & 3.15 & 2.85 & 2.36 & 3.80 & 2.08 & 3.05 \\
Direct distillation & 4.00 & 3.21 & 2.84 & 2.56 & 3.85 & 2.14 & 3.10 \\
Shift, $w{=}0.3$ & 3.94 & 3.17 & 2.98 & 2.60 & 3.79 & 2.16 & 3.11 \\
HAWK & 4.01 & 3.29 & 2.97 & 2.68 & 3.88 & 2.16 & \textbf{3.17} \\
\midrule
\multicolumn{8}{@{}l}{\emph{Speedup ratio}} \\
No auxiliary loss & 2.36 & 2.04 & 2.18 & 1.74 & 2.28 & 1.79 & 2.07 \\
Direct distillation & 2.32 & 2.02 & 2.19 & 1.80 & 2.25 & 1.69 & 2.05 \\
Shift, $w{=}0.3$ & 2.31 & 2.02 & 2.20 & 1.80 & 2.23 & 1.76 & 2.05 \\
HAWK & 2.33 & 2.06 & 2.22 & 1.88 & 2.27 & 1.78 & \textbf{2.09} \\
\bottomrule
\end{tabular}
\end{table}

\textbf{Layer-region fusion.}
Finally, we separate the effects of representation-based layer selection and learned fusion. Visual-state pooling and prediction-shift supervision are disabled in this study. As a reference, the EAGLE-3 layers $[1,14,26]$ under the same conditions give $\tau=3.05$ and $2.07\times$ (the no-pooling row of Table~\ref{tab:ablation-pooling}). Using uniformly spaced target layers with learned mixing gives an average acceptance length of 3.17 and a $2.09\times$ speedup. Replacing the uniform layer grid with CKA-selected layers while fixing the mixing weights to uniform averaging raises acceptance length to 3.21, with the same $2.09\times$ speedup. This comparison isolates the benefit of selecting layers according to representation similarity rather than position alone. Learning the mixing weights over the same CKA-selected layers further improves acceptance length to 3.24 and speedup to $2.11\times$ (Table~\ref{tab:ablation-fusion}).

\begin{table}[H]
\caption{Ablation of layer selection and fusion on SmolVLM-256M. The other components are disabled. Uniform layers are evenly spaced. CKA layers are selected by representation similarity. Learned mixing trains the region weights; uniform mixing freezes them at equal weights.}
\label{tab:ablation-fusion}
\centering
\footnotesize
\setlength{\tabcolsep}{3.5pt}
\renewcommand{\arraystretch}{1.0}
\begin{tabular}{@{}lrrrrrrr@{}}
\toprule
Setting & SEED-Bench & MM-Vet & MathVista & SQA & GQA & DocVQA & Avg. \\
\midrule
\multicolumn{8}{@{}l}{\emph{Acceptance length $\tau$}} \\
EAGLE-3 layers (reference) & 4.03 & 3.15 & 2.85 & 2.36 & 3.80 & 2.08 & 3.05 \\
Uniform layers, learned mix & 4.16 & 3.25 & 2.86 & 2.65 & 3.92 & 2.19 & 3.17 \\
CKA layers, uniform mix & 4.07 & 3.29 & 3.09 & 2.66 & 3.93 & 2.21 & 3.21 \\
CKA layers, learned mix & 4.18 & 3.34 & 3.14 & 2.67 & 3.90 & 2.22 & \textbf{3.24} \\
\midrule
\multicolumn{8}{@{}l}{\emph{Speedup ratio}} \\
EAGLE-3 layers (reference) & 2.36 & 2.04 & 2.18 & 1.74 & 2.28 & 1.79 & 2.07 \\
Uniform layers, learned mix & 2.38 & 2.06 & 2.21 & 1.85 & 2.28 & 1.75 & 2.09 \\
CKA layers, uniform mix & 2.33 & 2.07 & 2.27 & 1.85 & 2.26 & 1.78 & 2.09 \\
CKA layers, learned mix & 2.38 & 2.05 & 2.32 & 1.84 & 2.26 & 1.80 & \textbf{2.11} \\
\bottomrule
\end{tabular}
\end{table}

\begin{figure}[t]
\centering
\includegraphics[width=0.68\linewidth]{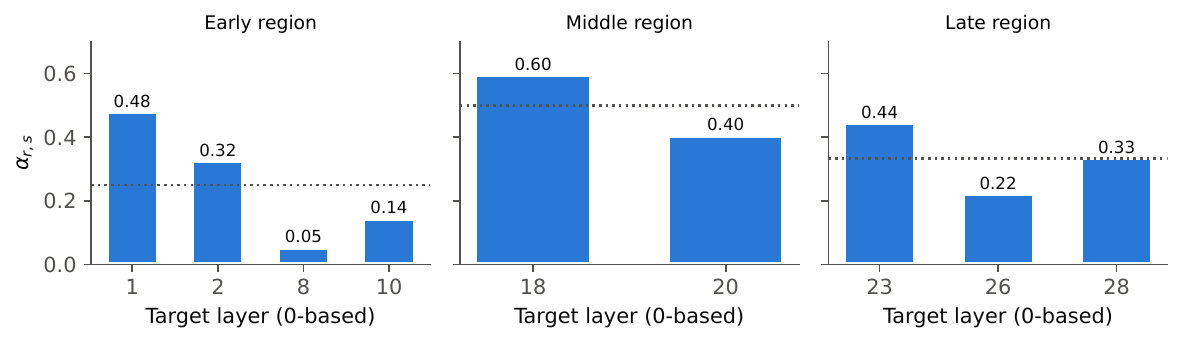}
\caption{Learned fusion weights $\alpha_{r,s}$ of the \alg{} drafter on
SmolVLM-256M. Dotted lines mark uniform weights.}
\label{fig:app-alpha}
\end{figure}

\subsection{Larger Target Models}
\label{sec:7b}

Table~\ref{tab:7b} reports results on LLaVA-1.6-7B and Qwen2.5-VL-7B. Due to
limited resources, both EAGLE-3 and \alg{} are trained on 30K examples for 10
epochs at this scale, and we do not retrain ViSpec or HiViS. Under greedy decoding, \alg{} raises average acceptance length from 3.73 to 4.02 on LLaVA-1.6-7B and from 3.19 to 3.40 on Qwen2.5-VL-7B, and average speedup from $2.37\times$ to $2.47\times$ and from $2.17\times$ to $2.25\times$. Both $\tau$ and SR improve on every benchmark at both temperatures.

\begin{table}[H]
\caption{Speedup ratio (SR) and average acceptance length ($\tau$) on LLaVA-1.6-7B and Qwen2.5-VL-7B under greedy decoding ($T=0$, top) and sampling ($T=1$, bottom). Both drafters are trained on 30K examples for 10 epochs.}
\label{tab:7b}
\centering
\renewcommand{\arraystretch}{1.2}
\begin{adjustbox}{max width=\linewidth}
\begin{tabular}{@{}l@{\hspace{4pt}}l@{\hspace{6pt}}r@{\hspace{4pt}}r@{\hspace{7pt}}r@{\hspace{4pt}}r@{\hspace{7pt}}r@{\hspace{4pt}}r@{\hspace{7pt}}r@{\hspace{4pt}}r@{\hspace{7pt}}r@{\hspace{4pt}}r@{\hspace{7pt}}r@{\hspace{4pt}}r@{\hspace{7pt}}r@{\hspace{4pt}}r@{\hspace{7pt}}r@{\hspace{4pt}}r@{\hspace{7pt}}r@{\hspace{4pt}}r@{\hspace{7pt}}r@{\hspace{4pt}}r@{\hspace{7pt}}r@{\hspace{4pt}}r@{}}
\toprule
 &  & \multicolumn{10}{c}{General} & \multicolumn{6}{c}{Reasoning} &
 \multicolumn{4}{c}{OCR} & \multicolumn{2}{c}{} \\
\cmidrule(lr){3-12} \cmidrule(lr){13-18} \cmidrule(lr){19-22}
\multirow{2}{*}{Model} & \multirow{2}{*}{Methods} & \multicolumn{2}{c}{MMMU} &
\multicolumn{2}{c}{SEED-Bench} & \multicolumn{2}{c}{MME} & \multicolumn{2}{c}{VQAv2} &
\multicolumn{2}{c}{MM-Vet} & \multicolumn{2}{c}{MathVista} &
\multicolumn{2}{c}{SQA} & \multicolumn{2}{c}{GQA} & \multicolumn{2}{c}{TextVQA}
& \multicolumn{2}{c}{DocVQA} & \multicolumn{2}{c}{Avg} \\
 & & SR & $\tau$ & SR & $\tau$ & SR & $\tau$ & SR & $\tau$ & SR & $\tau$ & SR &
 $\tau$ & SR & $\tau$ & SR & $\tau$ & SR & $\tau$ & SR & $\tau$ & SR & $\tau$ \\
\midrule
\multicolumn{24}{c}{\textbf{Greedy decoding ($T=0$)}} \\
\midrule
\multirow{2}{*}{\makecell[l]{LLaVA-1.6\\7B}}
 & EAGLE-3 & 2.32 & 3.59 & 2.57 & 4.30 & 2.29 & 3.51 & 2.66 & 4.53 & 2.33 & 3.62 & 2.34 & 3.65 & 2.18 & 3.20 & 2.58 & 4.31 & 2.28 & 3.49 & 2.15 & 3.13 & 2.37 & 3.73 \\
 & \alg{} & 2.41 & 3.84 & 2.73 & 4.74 & 2.37 & 3.73 & 2.79 & 4.92 & 2.43 & 3.89 & 2.43 & 3.90 & 2.28 & 3.49 & 2.68 & 4.60 & 2.39 & 3.79 & 2.21 & 3.28 & 2.47 & 4.02 \\
\midrule
\multirow{2}{*}{\makecell[l]{Qwen2.5-VL\\7B}}
 & EAGLE-3 & 1.99 & 2.67 & 2.39 & 3.79 & 2.11 & 3.00 & 2.48 & 4.05 & 2.16 & 3.15 & 2.10 & 2.99 & 2.07 & 2.89 & 2.41 & 3.85 & 2.06 & 2.87 & 1.96 & 2.60 & 2.17 & 3.19 \\
 & \alg{} & 2.10 & 2.98 & 2.51 & 4.12 & 2.17 & 3.18 & 2.62 & 4.43 & 2.22 & 3.32 & 2.16 & 3.15 & 2.12 & 3.03 & 2.49 & 4.07 & 2.11 & 3.00 & 2.00 & 2.71 & 2.25 & 3.40 \\
\midrule
\multicolumn{24}{c}{\textbf{Sampling ($T=1$)}} \\
\midrule
\multirow{2}{*}{\makecell[l]{LLaVA-1.6\\7B}}
 & EAGLE-3 & 1.83 & 2.68 & 2.19 & 3.55 & 1.90 & 2.86 & 2.26 & 3.71 & 1.95 & 2.96 & 1.86 & 2.76 & 1.83 & 2.69 & 2.19 & 3.54 & 1.81 & 2.64 & 1.74 & 2.45 & 1.96 & 2.98 \\
 & \alg{} & 1.92 & 2.90 & 2.23 & 3.64 & 2.00 & 3.08 & 2.38 & 4.02 & 2.02 & 3.15 & 1.93 & 2.92 & 1.93 & 2.93 & 2.28 & 3.77 & 1.88 & 2.79 & 1.81 & 2.63 & 2.04 & 3.18 \\
\midrule
\multirow{2}{*}{\makecell[l]{Qwen2.5-VL\\7B}}
 & EAGLE-3 & 1.74 & 2.47 & 2.16 & 3.47 & 1.87 & 2.78 & 2.18 & 3.52 & 1.93 & 2.91 & 1.85 & 2.72 & 1.86 & 2.74 & 2.07 & 3.26 & 1.75 & 2.48 & 1.68 & 2.31 & 1.91 & 2.87 \\
 & \alg{} & 1.82 & 2.65 & 2.23 & 3.64 & 1.92 & 2.90 & 2.27 & 3.74 & 1.96 & 2.99 & 1.94 & 2.95 & 1.89 & 2.82 & 2.16 & 3.49 & 1.78 & 2.57 & 1.72 & 2.41 & 1.97 & 3.02 \\
\bottomrule
\end{tabular}
\end{adjustbox}
\end{table}

\FloatBarrier
\section{Conclusion}
We introduced \alg{} to improve multimodal speculative decoding through better target conditioning and supervision. Layer-region fusion combines features from representation-based target regions, and visual-state pooling compresses target-processed visual states into a shorter sequence. Prediction-shift supervision teaches the drafter to match target logit shifts after plausible token substitutions. On SmolVLM-256M, \alg{} improves acceptance length and speedup over EAGLE-3, ViSpec, and HiViS under both greedy decoding and sampling, and on LLaVA-1.6-7B and Qwen2.5-VL-7B it improves both metrics over EAGLE-3 on every benchmark at both temperatures.

\subsection*{Reproducibility statement}
Appendix~\ref{app:impl} lists the drafter architecture, optimizer, learning-rate
schedule, training data, the target layers read by every reported drafter, and
the prediction-shift hyperparameters. Appendix~\ref{app:timing} gives the
hardware, decoding configuration, and metric definitions, and
Appendix~\ref{app:bench} the source, split, and prompt selection of every
benchmark. Appendix~\ref{app:cka} gives the CKA data, the region objective, and
the layer-selection algorithm with its selected layers. Appendix~\ref{app:baselines}
describes how the baselines were run.

\clearpage
\bibliography{iclr2027_conference}
\bibliographystyle{iclr2027_conference}
\clearpage
\appendix

\section{Implementation Details}
\label{app:impl}

\textbf{Layer indexing.} Target layers are indexed from 0 by decoder block:
layer $l$ is the output of the $l$-th decoder block, i.e.,
\texttt{hidden\_states[$l{+}1$]} in HuggingFace convention, where
\texttt{hidden\_states[0]} is the embedding output. SmolVLM-256M has $N=30$
decoder blocks, Qwen2.5-VL-7B has $N=28$, and LLaVA-1.6-7B has $N=32$. In this
indexing the EAGLE-3 rule selects blocks $(1,\,N/2-1,\,N-4)$, e.g., $[1,14,26]$
for SmolVLM-256M; this is $(2,\,N/2,\,N-3)$ in the 1-based notation
of~\citet{li2025eagle3}. The CKA search ranges over blocks $1$ to $N-2$.

\textbf{Drafter.} A single-layer EAGLE-3 drafter (hidden size 576, 9 attention
heads, 3 KV heads, intermediate size 1536) with a 32{,}000-token draft vocabulary
mapped from the 49{,}280-token target vocabulary, in bfloat16. Training-time
test unrolls 7 draft steps.

\textbf{Optimization.} AdamW ($\beta_1=0.9$, $\beta_2=0.999$, no weight decay),
constant learning rate $10^{-4}$ without warmup, gradient clipping at 1.0,
bfloat16, seed 42, 20 epochs, on 8 NVIDIA RTX A5000 GPUs.

\textbf{Training data.} Following HiViS~\citep{xie2026hivis}, we combine
68{,}000 text-only conversations from ShareGPT with 68{,}000 multimodal
conversations from LLaVA-Mix665K, filter out examples whose responses contain
fewer than five words, and regenerate assistant responses with the target
model, giving 132{,}929 training records. Target hidden states of the
selected layers and target logits are pre-computed once and stored as an
offline cache.

\textbf{Visual-state pooling.} SmolVLM encodes an image as 9, 13, or 17 tiles
depending on aspect ratio, with 64 image rows per tile (576, 832, or 1{,}088
rows; 832 for the 13-tile case quoted in the main text). Pooling acts on each
contiguous run of image rows independently. With $c=4$ each tile keeps 16 rows
(208 for 13 tiles); the $16\times$ setting keeps 4 rows per tile (52 for 13
tiles). Each pooled row takes the token id and position id of the first row in
its group, so text rows keep their original absolute positions and the RoPE
positions seen by the drafter are a subset of those seen by the target. Within a
tile, rows follow raster order over an $8\times8$ grid, so a group of four
consecutive rows covers a $1\times4$ horizontal strip of patches. The EAGLE-3
baseline receives all image rows.

\textbf{Target layers.} Table~\ref{tab:app-layers} lists the target layers read
by each SmolVLM-256M drafter.

\begin{table}[h]
\caption{Target layers (0-based decoder blocks) read by each SmolVLM-256M drafter.}
\label{tab:app-layers}
\centering
\small
\begin{tabular}{@{}lll@{}}
\toprule
Drafter & Early / Middle / Late & Used in \\
\midrule
EAGLE-3 & $[1]$ / $[14]$ / $[26]$ & Table~\ref{tab:main-results}; Tables~\ref{tab:ablation-pooling}--\ref{tab:ablation-shift} \\
\alg{} & $[1,2,8,10]$ / $[18,20]$ / $[23,26,28]$ & Table~\ref{tab:main-results} \\
CKA coverage, $\epsilon_{\mathrm{CKA}}=0.008$ & $[1,2,5,9]$ / $[14,18,21]$ / $[23,25,27]$ & Table~\ref{tab:ablation-fusion} \\
Uniform grid & $[1,4,8,11]$ / $[14,18]$ / $[21,24,28]$ & Table~\ref{tab:ablation-fusion} \\
\bottomrule
\end{tabular}
\end{table}

\textbf{Prediction-shift supervision.} $\lambda_{\mathrm{shift}}=0.1$, drafter
proposal $=$ drafter top-1, substitution at the first rollout step only.

\section{Hardware and Timing Protocol}
\label{app:timing}

All evaluation runs on a single NVIDIA RTX A5000 (24\,GB) with the other GPUs on
the machine idle, one prompt at a time (batch size 1), in bfloat16, with three
warmup prompts before timing. Tree decoding uses depth 5, 60 candidate tokens,
and top-$k$ 10. Prompts are limited to 4{,}000 input tokens and generation to
500 new tokens. At $T=1$, the target samples from its full output distribution
without top-$k$ or nucleus truncation. Acceptance length
$\tau$ is, per prompt, the mean over verification rounds of the number of
accepted draft tokens plus one, averaged over prompts. Speedup ratio follows the
definition of HiViS~\citep{xie2026hivis}: per prompt, the method's generation
token rate divided by that of autoregressive decoding, averaged over prompts.
Greedy results are reported on prompts whose decoded answer is identical to the
autoregressive answer for every method in the table.

\section{Baseline Reproduction}
\label{app:baselines}

EAGLE-3 uses the same target, training data, offline cache, optimizer, and
number of epochs as \alg{}, reads the EAGLE-3 layers in
Table~\ref{tab:app-layers}, and receives all image rows. ViSpec and HiViS are
trained with their official implementations. All four methods are evaluated
with the same evaluator, decoding loop, tree configuration, prompts, and prompt
filter, each against an autoregressive baseline run on the same machine.

\section{CKA Layer Selection}
\label{app:cka}

\textbf{Data.} We run each target on 48 training conversations and pool
representations over all text positions (6{,}438 for SmolVLM-256M, 11{,}193 for
Qwen2.5-VL-7B, 10{,}177 for LLaVA-1.6-7B). Image positions are excluded: the
drafting loss is applied at text positions, and hundreds of near-duplicate
visual rows would otherwise dominate the statistic.

\textbf{Outlier dimensions.} From block 11 onward in SmolVLM-256M, a single
hidden dimension carries most of the variance, and with it included linear CKA
between any two adjacent layers in blocks 11--28 is 1.000. We therefore remove
the 32 highest-variance dimensions, chosen once over all layers so that every
layer is compared in the same subspace, before computing CKA.

\textbf{Region boundaries.} Let $\mathcal{L}=\{1,\dots,N-2\}$. We search over
all partitions of $\mathcal{L}$ into three contiguous regions
$\mathcal{R}_1,\mathcal{R}_2,\mathcal{R}_3$ of at least two layers each and
choose
\begin{equation}
(\mathcal{R}_1,\mathcal{R}_2,\mathcal{R}_3)
=\arg\max\;
\underset{\substack{r;\;i<j\\ i,j\in\mathcal{R}_r}}{\mathrm{mean}}\,\mathrm{CKA}(i,j)
\;-\;
\underset{\substack{r<r'\\ i\in\mathcal{R}_r,\,j\in\mathcal{R}_{r'}}}{\mathrm{mean}}\,\mathrm{CKA}(i,j),
\end{equation}
by exhaustive search over the $O(N^2)$ boundary pairs.

\begin{figure}[h]
    \centering
    \begin{subfigure}[t]{0.82\linewidth}
        \centering
        \includegraphics[width=\linewidth]
            {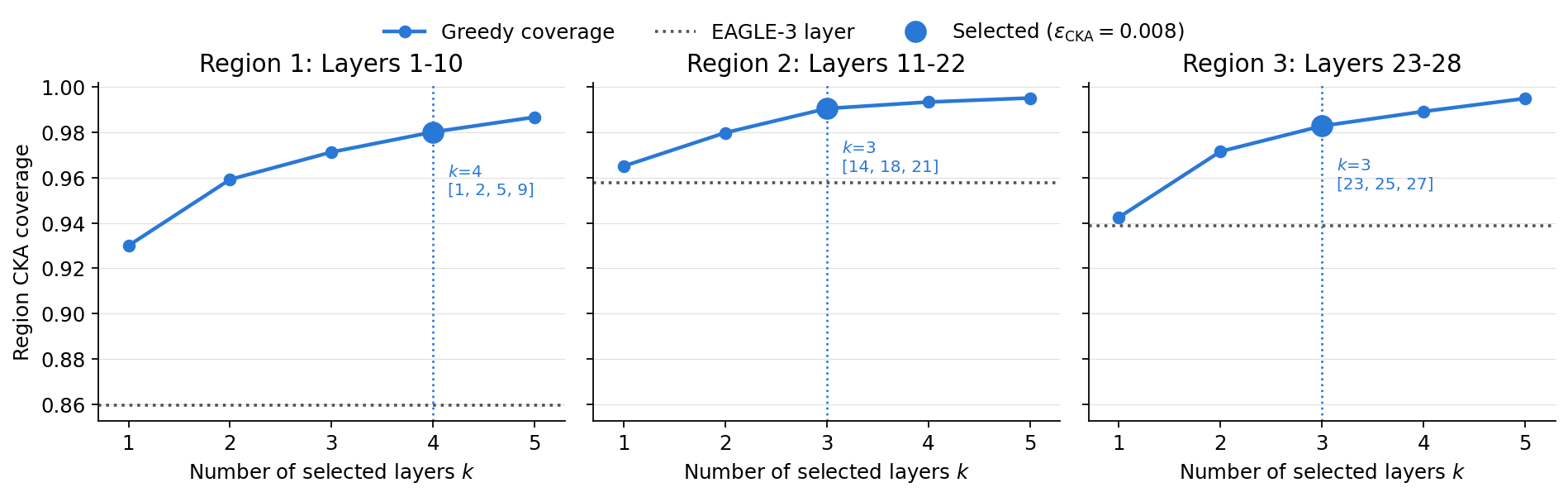}
        \caption{SmolVLM-256M.}
        \label{fig:layer-region-coverage-smolvlm}
    \end{subfigure}

    \vspace{2mm}

    \begin{subfigure}[t]{0.82\linewidth}
        \centering
        \includegraphics[width=\linewidth]
            {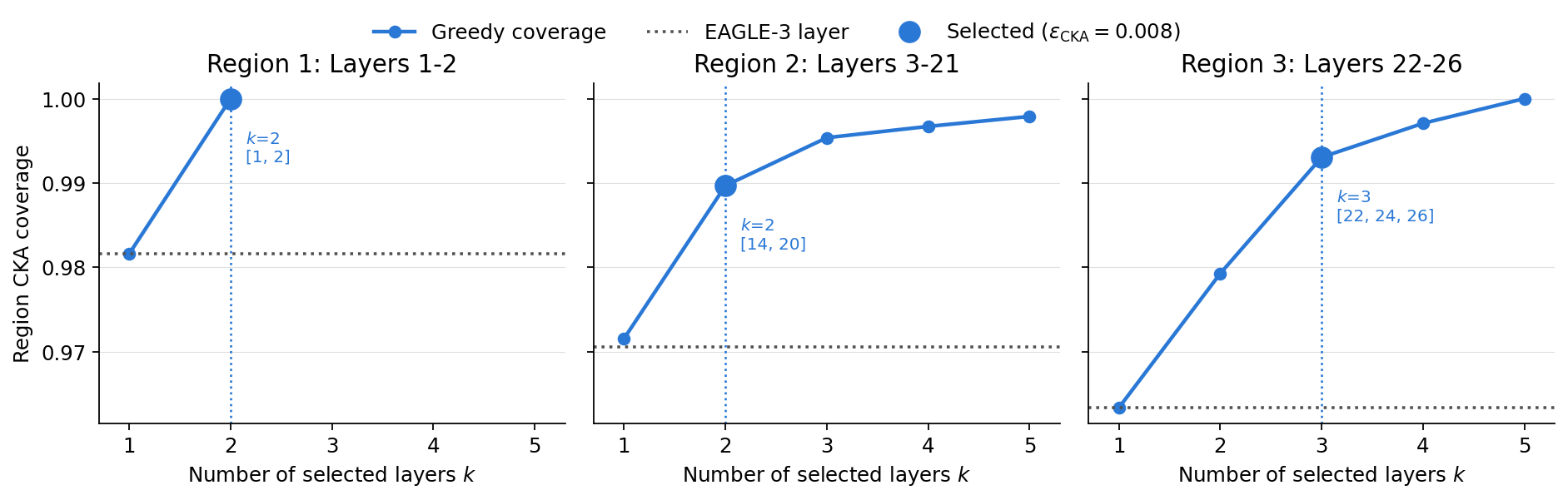}
        \caption{Qwen2.5-VL-7B.}
        \label{fig:layer-region-coverage-qwen}
    \end{subfigure}

    \caption{Region-wise CKA coverage as the number of selected target layers
    increases for (a) SmolVLM-256M and (b) Qwen2.5-VL-7B. Large markers show
    where the greedy selection stops at $\epsilon_{\mathrm{CKA}}=0.008$; dotted
    lines show the coverage of the single EAGLE-3 layer in each region.}
    \label{fig:layer-region-coverage}
\end{figure}

\textbf{Layer selection.} Within each region, Algorithm~1 greedily adds the
layer that most increases $C_{\mathrm{CKA}}$ until the gain falls to
$\epsilon_{\mathrm{CKA}}=0.008$ or below. The first layer added is the region's
coverage medoid. Table~\ref{tab:app-cka} lists the regions and selected layers
of each target, and Figure~\ref{fig:layer-region-coverage} shows how coverage
grows as layers are added.

\begin{center}
\begin{minipage}{0.9\linewidth}
\small
\hrule\vspace{3pt}
\textbf{Algorithm 1} Greedy CKA-coverage selection for region $\mathcal{R}$
\vspace{2pt}\hrule\vspace{3pt}
\begin{tabbing}
\hspace{1em}\=\hspace{1em}\=\kill
$\mathcal{S}\leftarrow\emptyset$\\
\textbf{while} $\mathcal{R}\setminus\mathcal{S}\neq\emptyset$:\\
\> $s^\star\leftarrow\arg\max_{s\in\mathcal{R}\setminus\mathcal{S}} C_{\mathrm{CKA}}(\mathcal{S}\cup\{s\})$\\
\> $\delta\leftarrow C_{\mathrm{CKA}}(\mathcal{S}\cup\{s^\star\})-C_{\mathrm{CKA}}(\mathcal{S})$\\
\> \textbf{if} $\mathcal{S}\neq\emptyset$ \textbf{and} $\delta\le\epsilon_{\mathrm{CKA}}$: \textbf{break}\\
\> $\mathcal{S}\leftarrow\mathcal{S}\cup\{s^\star\}$\\
\textbf{return} $\mathcal{S}$
\end{tabbing}
\vspace{-4pt}\hrule
\end{minipage}
\end{center}

\begin{table}[h]
\caption{CKA regions and coverage selections (0-based blocks, $\epsilon_{\mathrm{CKA}}=0.008$). Layers are listed in order of addition; parentheses give each layer's coverage gain.}
\label{tab:app-cka}
\centering
\small
\begin{adjustbox}{max width=\linewidth}
\begin{tabular}{@{}llll@{}}
\toprule
Target & Early & Middle & Late \\
\midrule
SmolVLM-256M & 1--10: 5, 9, 2, 1 & 11--22: 18, 21, 14 & 23--28: 25, 27, 23 \\
 & (.930, .029, .012, .009) & (.965, .015, .011) & (.943, .029, .011) \\
Qwen2.5-VL-7B & 1--2: 1, 2 & 3--21: 14, 20 & 22--26: 24, 26, 22 \\
 & (.982, .018) & (.972, .018) & (.963, .016, .014) \\
LLaVA-1.6-7B & 1--16: 10, 15 & 17--18: 17, 18 & 19--30: 25, 20, 28 \\
 & (.980, .011) & (.990, .010) & (.964, .016, .009) \\
\bottomrule
\end{tabular}
\end{adjustbox}
\end{table}

\textbf{Qwen2.5-VL-7B Region 1.} The early region of Qwen2.5-VL-7B contains
only blocks 1--2 because the transition from block 2 to block 3 is the sharpest
in the network: CKA between blocks 2 and 3 is 0.10, while every adjacent pair
from block 3 to block 26 has CKA of at least 0.95. The mean CKA between blocks
1--2 and blocks 3--21 is 0.22 (Figure~\ref{fig:layer-band-zone-qwen}).

\section{Learned Fusion Weights}
\label{app:alpha}

Figure~\ref{fig:app-alpha} shows the softmax-normalized fusion weights
$\alpha_{r,s}$ of the \alg{} drafter after training. The weights depart from
uniform in every region: blocks 1 and 2 receive 0.48 and 0.32 of the early
region while block 8 receives 0.05; block 18 receives 0.60 of the middle region;
and block 23 receives 0.44 of the late region.

\section{Visual Compression Diagnostic}
\label{app:compression}

\begin{figure}[h]
\centering
\includegraphics[width=0.8\linewidth]{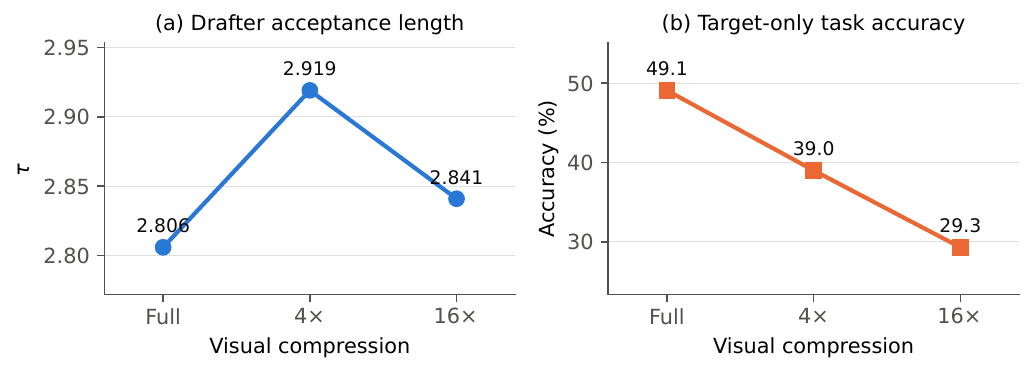}
\caption{Visual compression diagnostic. (a) Acceptance length of drafters
conditioned on pooled target visual states. (b) Task accuracy of the target when
its own visual input is compressed by the same factor.}
\label{fig:vhsc-compression}
\end{figure}

Figure~\ref{fig:vhsc-compression} relates the compression factor of the visual
hidden states to drafting and to the target's own answers. Panel (a) reports the
acceptance length of drafters conditioned on pooled target visual states:
$4\times$ pooling gives the highest acceptance length, and $16\times$ pooling
falls back toward the uncompressed setting. Panel (b) is a separate target-only
experiment in which the target itself receives the compressed visual input; its
accuracy drops from 49.1 to 39.0 at $4\times$ and to 29.3 at $16\times$. Panel
(b) only measures how much information compression removes. In every
speculative-decoding result in this paper the target receives the full visual
input, so compression changes what the drafter sees but never the target's
output.

\section{Prediction-Shift Supervision: Details}
\label{app:shift}

\textbf{The shift is a centered log-probability ratio.} For logits $z$ and
$p=\mathrm{softmax}(z)$, $\log p_i=z_i-\log\sum_j e^{z_j}$, so
$\mathrm{center}(z)_i=\log p_i-\frac{1}{|V|}\sum_j\log p_j$. Hence
\begin{equation}
\Delta_{T,t,i}
=\log\frac{p_T^{\mathrm{alt}}(i)}{p_T^{\mathrm{orig}}(i)}
-\frac{1}{|V|}\sum_{j}\log\frac{p_T^{\mathrm{alt}}(j)}{p_T^{\mathrm{orig}}(j)},
\end{equation}
and likewise for $\Delta_{D,t}$. $\mathcal{L}_{\mathrm{shift}}=0$ exactly when
$p_D^{\mathrm{alt}}(i)/p_D^{\mathrm{orig}}(i)\propto
p_T^{\mathrm{alt}}(i)/p_T^{\mathrm{orig}}(i)$ for every token $i$. Combined
with $\mathcal{L}_{\mathrm{std}}$, which fits $p_D^{\mathrm{orig}}$ to
$p_T^{\mathrm{orig}}$, this fits $p_D^{\mathrm{alt}}$ to $p_T^{\mathrm{alt}}$
through the ratio rather than through the level: $\mathcal{L}_{\mathrm{std}}$
supervises the prediction, and $\mathcal{L}_{\mathrm{shift}}$ supervises how the
prediction changes when the input token changes. As a function of the input
token, $\Delta$ is a finite difference of the log-probabilities along the
direction $x_t\to\hat{x}_t$. Unlike cross-entropy on $p_T^{\mathrm{alt}}$, the
squared error on centered logits weights every draft-vocabulary entry equally
rather than concentrating on high-probability tokens. The squared error is
averaged over the draft vocabulary.

\textbf{Relation to training-time test.} In EAGLE-3's training-time test, the
drafter at rollout step $k>1$ receives its own output feature from step $k-1$,
but its input token is always the training-sequence token $x_t$. It is
therefore trained on its own feature drift but never on a context whose token
differs from the training sequence. Prediction-shift supervision runs at exactly
the training-time-test state of the second rollout step: the same drafter
feature (its own output from the first step), the same cache over the prefix
(built from target features), and only the input token replaced by $\hat{x}_t$.
It thus supervises the token pathway that training-time test leaves
unsupervised. HASS~\citep{zhang2025hass} likewise aligns training with the
drafter's own features, and DistillSpec~\citep{zhou2024distillspec} distills
the target distribution on drafter-generated contexts, which corresponds to our
direct-distillation control.

\textbf{Why $\hat{x}_t\neq$ target top-1.} Responses
are generated by the target, so $x_t$ is usually the target's top-1 token, and
substituting it would leave the context unchanged.

\textbf{Target forward.} The offline cache holds the target's hidden states and
logits on the original sequence, which give $z_{T}^{\mathrm{orig}}$. The
substituted sequence $\bar{x}$ is scored by a live target forward during
training, one sample at a time over its unpadded length, which gives
$z_{T}^{\mathrm{alt}}$. No hidden state from this forward reaches the drafter.

\begin{center}
\begin{minipage}{0.92\linewidth}
\small
\hrule\vspace{3pt}
\textbf{Algorithm 2} One training step with prediction-shift supervision
\vspace{2pt}\hrule\vspace{3pt}
\begin{tabbing}
\hspace{1em}\=\hspace{1em}\=\kill
Read $(x,\,\mathbf{H},\,z_T^{\mathrm{orig}})$ from the offline cache\\
Rollout step 1: drafter logits $z_D^{(1)}$; \ $\hat{x}_t\leftarrow\arg\max z_{D,t}^{(1)}$\\
$\mathcal{S}\leftarrow\{t:\hat{x}_t\ne\arg\max z_{T,t}^{\mathrm{orig}}\}$\\
$\bar{x}\leftarrow x$ with $\bar{x}_t=\hat{x}_t$ for $t\in\mathcal{S}$; \ $z_T^{\mathrm{alt}}\leftarrow\mathrm{Target}(\bar{x})$ \quad (no gradient)\\
Rollout step 2 (normal): input tokens $x$ $\to$ $z_D^{\mathrm{orig}}$\\
Rollout step 2 (branch): same features and cache, tokens $\hat{x}$ at $\mathcal{S}$ $\to$ $z_D^{\mathrm{alt}}$\\
$\mathcal{L}_{\mathrm{shift}}\leftarrow\mathrm{mean}_{t\in\mathcal{S}}\,\frac{1}{|V_D|}\|\Delta_{D,t}-\Delta_{T,t}\|_2^2$\\
$\mathcal{L}\leftarrow\mathcal{L}_{\mathrm{std}}+\lambda_{\mathrm{shift}}\mathcal{L}_{\mathrm{shift}}$
\end{tabbing}
\vspace{-4pt}\hrule
\end{minipage}
\end{center}

\section{Benchmark Details}
\label{app:bench}

Table~\ref{tab:app-bench} lists the source and split of each benchmark. For
benchmarks marked ``shuffle'', the 80 prompts are the first 80 after shuffling
with seed 42; TextVQA and VQAv2 take the first 80 records in file order.

\begin{table}[h]
\caption{Benchmark sources. 80 prompts each, at most 500 generated tokens.}
\label{tab:app-bench}
\centering
\small
\begin{adjustbox}{max width=\linewidth}
\begin{tabular}{@{}llll@{}}
\toprule
Benchmark & Source & Split & Selection \\
\midrule
MMMU & \texttt{MMMU/MMMU} (Accounting, Art, Biology, Math) & test & shuffle; $\le3$ images \\
SEED-Bench & LLaVA SEED-Bench question file & -- & shuffle \\
MME & LLaVA MME question file & -- & shuffle \\
VQAv2 & \texttt{lmms-lab-encoder/VQAv2} & test & first 80 \\
MM-Vet & \texttt{mm-vet.json} & -- & shuffle \\
MathVista & \texttt{AI4Math/MathVista} & testmini & shuffle; image, English only \\
ScienceQA & \texttt{derek-thomas/ScienceQA} & validation & shuffle; image only \\
GQA & LLaVA GQA question file & testdev-balanced & shuffle \\
TextVQA & TextVQA 0.5.1 & test & first 80 \\
DocVQA & \texttt{lmms-lab-encoder/DocVQA} & test & shuffle \\
\bottomrule
\end{tabular}
\end{adjustbox}
\end{table}

\section{Acceptance by Draft Step}
\label{app:per-step}

Table~\ref{tab:app-per-step} and Figure~\ref{fig:app-per-step} break acceptance
down by draft step, computed from the evaluation logs behind
Tables~\ref{tab:main-results} and~\ref{tab:ablation-shift}. For every
verification round, step $k$ counts as accepted when at least $k$ draft tokens
are accepted; values are pooled over rounds within a benchmark and averaged over
the ten benchmarks.

\begin{table}[H]
\caption{$P(\text{accepted length}\ge k)$ by draft step $k$, greedy decoding,
ten benchmarks. 20-epoch rows use prompts whose output matches autoregressive
decoding; 4-epoch rows use all prompts.}
\label{tab:app-per-step}
\centering
\small
\begin{tabular}{@{}lrrrrr@{}}
\toprule
 & $k=1$ & $k=2$ & $k=3$ & $k=4$ & $k=5$ \\
\midrule
EAGLE-3 (20 ep) & 0.789 & 0.603 & 0.363 & 0.248 & 0.161 \\
\alg{} (20 ep) & 0.840 & 0.702 & 0.525 & 0.365 & 0.233 \\
\midrule
No auxiliary loss (4 ep) & 0.767 & 0.585 & 0.344 & 0.214 & 0.136 \\
Direct distillation (4 ep) & 0.770 & 0.595 & 0.354 & 0.235 & 0.148 \\
Prediction shift (4 ep) & 0.789 & 0.612 & 0.373 & 0.248 & 0.168 \\
\bottomrule
\end{tabular}
\end{table}

\end{document}